\documentclass[a4paper,10pt]{llncs}
\usepackage[utf8]{inputenc}
\usepackage{graphicx}
\usepackage{comment}
\usepackage{booktabs}
\usepackage{tabu}

\begin{document}

\author{Mohsen Ghafoorian\inst{1,2,3}
\and Alireza Mehrtash\inst{2,4}
\thanks{
Mohsen Ghafoorian and Alireza Mehrtash contributed equally to this work.
}
\and Tina Kapur\inst{2} 
\and Nico Karssemeijer\inst{1} 
\and Elena Marchiori\inst{3}
\and Mehran Pesteie\inst{4} 
\and Charles R. G. Guttmann\inst{2}
\and Frank-Erik de Leeuw\inst{5} 
\and Clare M. Tempany\inst{2} 
\and Bram van Ginneken\inst{1}
\and Andriy Fedorov\inst{2} 
\and Purang Abolmaesumi\inst{4} 
\and Bram Platel\inst{1} 
\and William M. Wells III\inst{2}
}

\institute{
Diagnostic Image Analysis Group, Radboud University Medical Center, Nijmegen, the Netherlands\\
\email{mohsen.ghafoorian@radboudumc.nl}
\and Radiology Department, Brigham and Women's Hospital, Harvard Medical School, Boston, MA, USA\\
\email{mehrtash@bwh.harvard.edu}
\and Institute for Computing and Information Sciences, Radboud University, Nijmegen, the Netherlands
\and University of British Columbia, Vancouver, BC, Canada
\and Donders Institute for Brain, Cognition and Behaviour, Department of Neurology, Radboud University Medical Center, Nijmegen, the Netherlands
}

\title{Transfer Learning for Domain Adaptation in MRI: Application in Brain Lesion Segmentation}
\titlerunning{Transfer Learning for Domain Adaptation to Brain MRI Segmentation}

\maketitle

\begin{abstract}
Magnetic Resonance Imaging (MRI) is widely used in routine clinical diagnosis and treatment. 
However, variations in MRI acquisition protocols result in different appearances of normal and diseased tissue in the images. 
Convolutional neural networks (CNNs), which have shown to be successful in many medical image analysis tasks, are typically sensitive to the variations in imaging protocols. 
Therefore, in many cases, networks trained on data acquired with one MRI protocol, do not perform satisfactorily on data acquired with different protocols. 
This limits the use of models trained with large annotated legacy datasets on a new dataset with a different domain which is often a recurring situation in clinical settings. In this study, we aim to answer the following central questions regarding domain adaptation in medical image analysis: Given a fitted legacy model, 1) How much data from the new domain is required for a decent adaptation of the original network?; and,
2) What portion of the pre-trained model parameters should be retrained given a certain number of the new domain training samples?
To address these questions, we conducted extensive experiments in white matter hyperintensity segmentation task. 
We trained a CNN on legacy MR images of brain and evaluated the performance of the domain-adapted network on the same task with images from a different domain. 
We then compared the performance of the model to the surrogate scenarios where either the same trained network is used or a new network is trained from scratch on the new dataset.
The domain-adapted network tuned only by two training examples achieved a Dice score of 0.63 substantially outperforming a similar network trained on the same set of examples from scratch.
\end{abstract}

\section{Introduction}
Deep neural networks have been extensively used in medical image analysis and have outperformed the conventional methods for specific tasks such as segmentation, classification and detection \cite{litjens2017}. For instance on brain MR analysis, convolutional neural networks (CNN) have been shown to achieve outstanding performance for various tasks including white matter hyperintensities (WMH) segmentation \cite{ghafoorian2016location}, tumor segmentation \cite{kamnitsas2017efficient}, microbleed detection \cite{dou2016automatic}, and lacune detection \cite{Ghafoorian_2017}.
Although many studies report excellent results on specific domains and image acquisition protocols, the generalizability of these models on test data with different distributions are often not investigated and evaluated.
Therefore, to ensure the usability of the trained models in real world practice, which involves imaging data from various scanners and protocols, domain adaptation remains a valuable field of study.
This becomes even more important when dealing with Magnetic Resonance Imaging (MRI), which demonstrates high variations in soft tissue appearances and contrasts among different protocols and settings.

Mathematically, a domain $D$ can be expressed by a feature space $\scalebox{1.35}{\raisebox{1pt}{$\chi$}}$ and a marginal probability distribution $P(X)$, where $X = \{x_1,...,x_n\}\in \scalebox{1.35}{\raisebox{1pt}{$\chi$}}$ \cite{pan2010survey}. 
A supervised learning task on a specific domain $D = \{\scalebox{1.35}{\raisebox{1pt}{$\chi$}} , P(X)\}$, consists of a pair of a label space $Y$ and an objective predictive function $f(.)$ (denoted by $T  = \{Y, f(.)\}$).
The objective function $f(.)$ can be learned from the training data, which consists of pairs $\{x_i, y_i\}$, where $x_i \in X$ and $y_i \in Y$. 
After the training process, the learned model denoted by $\tilde{f}(.)$ is used to predict the label for a new instance $x$.
Given a source domain $D_S$ with a learning task $T_S$ and a target domain $D_T$ with learning task $T_T$, transfer learning is defined as the process of improving the learning of the target predictive function $f_T(.)$ in $D_T$ using the information in 
$D_S$ and $T_S$, where $D_S\neq D_T$, or $T_S\neq T_T$ \cite{pan2010survey}. 
We denote $\tilde{f}_{ST}(.)$ as the predictive model initially trained on the source domain $D_S$, and domain-adapted to the target domain $D_T$.


In the medical image analysis literature, transfer classifiers such as adaptive SVM and transfer AdaBoost, are shown to outperform the common supervised learning approaches in segmenting brain MRI, trained only on a small set of target domain images \cite{van2015transfer}. 
In another study a machine learning based sample weighting strategy was shown to be capable of handling multi-center chronic obstructive pulmonary disease images \cite{cheplygina2017transfer}.
Recently, also several studies have investigated transfer learning methodologies on deep neural networks applied to medical image analysis tasks. A number of studies used networks pre-trained on natural images to extract features and followed by another classifier, such as a Support Vector Machine (SVM) or a random forest~\cite{esteva2017dermatologist}. Other studies~\cite{tajbakhsh2016convolutional,Shin2016} performed layer fine-tuning on the pre-trained networks for adapting the learned features to the target domain.

Considering the hierarchical feature learning fashion in CNN, we expect the first few layers to learn features for general simple visual building blocks, such as edges, corners and simple blob-like structures, while the deeper layers learn more complicated abstract task-dependent features. In general, the ability to learn domain-dependent high-level representations is an advantage enabling CNNs to achieve great recognition capabilities. 
However, it is not obvious how these qualities are preserved during the transfer learning process for domain adaptation. 
For example, it would be practically important to determine how much data on the target domain is required for domain adaptation with sufficient accuracy for a given task, or how many layers from a model fitted on the source domain can be effectively transferred to the target domain. 
Or more interestingly, given a number of available samples on the target domain, what layer types and how many of those can we afford to fine-tune. 
Moreover, there is a common scenario in which a large set of annotated legacy data is available, often collected in a time-consuming and costly process. 
Upgrades in the scanners, acquisition protocols, etc., as we will show, might make the direct application of models trained on the legacy data unsuccessful. 
To what extent these legacy data can contribute to a better analysis of new datasets, or vice versa, is another question worth investigating.

In this study, we aim towards answering the questions discussed above. We use transfer learning methodology for domain adaptation of models trained on legacy MRI data on brain WMH segmentation. 

\section{Materials and Method}
\subsection{Dataset}
Radboud University Nijmegen Diffusion tensor and Magnetic resonance imaging Cohort (RUN DMC) \cite{Norden11c} is a longitudinal study of patients diagnosed with small vessel disease. 
The baseline scans acquired in 2006 consisted of fluid-attenuated inversion recovery (FLAIR) images with voxel size of 1.0$\times$1.2$\times$5.0 mm and an inter-slice gap of 1.0 mm, scanned with a 1.5~T Siemens scanner. 
However, the follow-up scans in 2011 were acquired differently with a voxel size of 1.0$\times$1.2$\times$3.0 mm, including a slice gap of 0.5 mm. 
The follow-up scans demonstrate a higher contrast as the partial volume effect is less of an issue due to thinner slices. 
For each subject, we also used 3D T1 magnetization-prepared rapid gradient-echo (MPRAGE) with voxel size of 1.0$\times$1.0$\times$1.0 mm which is the same among the two datasets. 
Reference WMH annotations on both datasets were provided semi-automatically, by manually editing segmentations provided by a WMH segmentation method \cite{ghafoorian2016automated} wherever needed.

The T1 images were linearly registered to FLAIR scans, followed by brain extraction and bias-filed correction operations. We then normalized the image intensities to be within the range of [0, 1].

In this study, we used 280 patient acquisitions with WMH annotations from the baseline as the source domain, and 159 scans from all the patients that were rescanned in the follow-up as the target domain.
Table~\ref{tab:sets} shows the data split into the training, validation and test sets. It should be noted that the same patient-level partitioning which was used on the baseline, was respected on the follow-up dataset to prevent potential label leakages.

\begin{table}
\centering
\caption{Number of patients for the domain adaptation experiments.}\label{tab:sets}
\label{table}
\begin{tabular}{@{}lcccccc@{}}
\toprule
    & \multicolumn{3}{c}{Source Domain} & \multicolumn{3}{c}{Target Domain} \\ \midrule
Set    & Train  & Validation & Test & Train  & Validation & Test \\
Size   & 200    & 30         & 50   & 100    & 26         & 33   \\ \bottomrule
\end{tabular}
\end{table}

\subsection{Sampling}
We sampled 32$\times$32 patches to capture local neighborhoods around WMH and normal voxels from both FLAIR and T1 images. We assigned each patch with the label of the corresponding central voxel. To be more precise, we randomly selected 25\% of all voxels within the WMH masks, and randomly selected the same number of negative samples from the normal appearing voxels inside the brain mask. We augmented the dataset by flipping the patches along the $y$ axis. This procedure resulted in training and validation datasets of size $\sim$1.2m and $\sim$150k on the baseline, and $\sim$1.75m and $\sim$200k on the followup.
\subsection{Network Architecture and Training}
\label{subsec:training}
\begin{figure}[t]
\centering
\vspace{1cm}
{\includegraphics[scale=0.49]{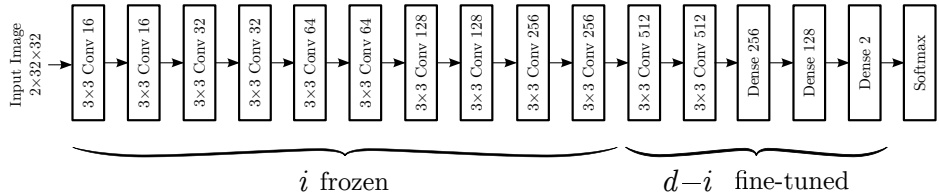}}
\caption{Arcitecture of the convolutional neural network used in our experiments. The shallowest $i$ layers are frozen and the rest $d-i$ layers are fine-tuned. $d$ is the depth of the network which was 15 in our experiments.}
\label{fig:arch}
\end{figure}

We stacked the FLAIR and T1 patches as the input channels and used a 15-layer architecture consisting of 12 convolutional layers of 3$\times$3 filters and 3 dense layers of 256, 128 and 2 neurons, and a final softmax layer. We avoided using pooling layers as they would result in a shift-invariance property that is not desirable in segmentation tasks, where the spatial information of the features are important to be preserved. The network architecture is illustrated in Figure~\ref{fig:arch}.

To tune the weights in the network, we used the Adam update rule~\cite{kingma2014adam} with a mini-batch size of 128 and a binary cross-entropy loss function. 
We used the Rectified Linear Unit (ReLU) activation function as the non-linearity and the He method \cite{he2015delving} that randomly initializes the weights drawn from a $\mathcal{N}(0, \sqrt{\frac{2}{m}})$  distribution, where $m$ is the number of inputs to a neuron. Activations of all layers were batch-normalized to speed up the convergence \cite{ioffe2015batch}.
A decaying learning rate was used with a starting value of $0.0001$ for the optimization process.
To avoid over-fitting, we regularized our networks with a drop-out rate of 0.3 as well as the $L_2$ weight decay with $\lambda_2$=0.0001. We trained our networks for a maximum of 100 epochs with an early stopping policy. For each experiment, we picked the model with the highest area under the curve on the validation set.

We trained our networks with a patch-based approach. 
At segmentation time, however, we converted the dense layers to their equivalent convolutional counterparts to form a fully convolutional network (FCN). 
FCNs are much more efficient as they avoid the repetitive computations on neighboring patches by feeding the whole image into the network. 
We prefer the conceptual distinction between dense and convolutional layers at the training time, to keep the generality of experiments for classification problems as well (e.g., testing the benefits of fine-tuning the convolutional layers in addition to the dense layers).
Patch-based training allows class-specific data augmentation to handle domains with hugely imbalanced class ratios (e.g., WMH segmentation domain).

\subsection{Domain Adaptation}

To build the model $\tilde{f}_{ST}(.)$, we transferred the learned weights from $\tilde{f}_S$, then we froze shallowest $i$ layers and fine-tuned the remaining $d-i$ deeper layers with the training data from $D_T$, where $d$ is the depth of the trained CNN. This is illustrated in Figure~\ref{fig:arch}.
We used the same optimization update-rule, loss function, and regularization techniques as described in Section~\ref{subsec:training}.

\subsection{Experiments}
On the WMH segmentation domain, we investigated and compared three different scenarios: 
1) Training a model on the source domain and directly applying it on the target domain; 
2) Training networks on the target domain data from scratch; 
and 3) Transferring model learned on the source domain onto the target domain with fine-tuning. 
In order to identify the target domain dataset sizes where transfer learning is most useful, the second and third scenarios were explored with different training set sizes of 2, 3, 4, 5, 6, 7, 8, 9, 10, 11, 12, 25, 50 and 100 cases. 
We extensively expanded the third scenario investigating the best freezing/tuning cut-off for each of the mentioned target domain training set sizes. 
We used the same network architecture and training procedure among the different experiments. 
The reported metric for the segmentation quality assessment is the Dice score.
\begin{figure}[!t]
\centering
\vspace{1cm}
{\includegraphics[scale=0.5]{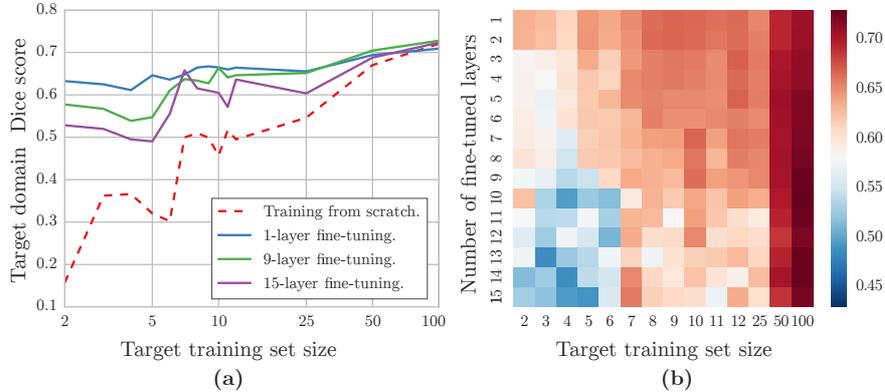}}
\caption{\textbf{(a)} The comparison of Dice scores on the target domain with and without transfer learning. A logarithmic scale is used on the $x$ axis.
\textbf{(b)} Given a deep CNN with $d$=15 layers, transfer learning was performed by freezing the $i$ initial layers and fine-tuning the last $d-i$ layers. The Dice scores on the test set are illustrated with the color-coded heatmap. On the map, the number of fine-tuned layers are shown horizontally, whereas the target domain training set size is shown vertically.}
\label{fig:plots}
\end{figure}

\section{Results}
\begin{figure}[!t]
\centering
\vspace{1cm}
{\includegraphics[scale=0.5]{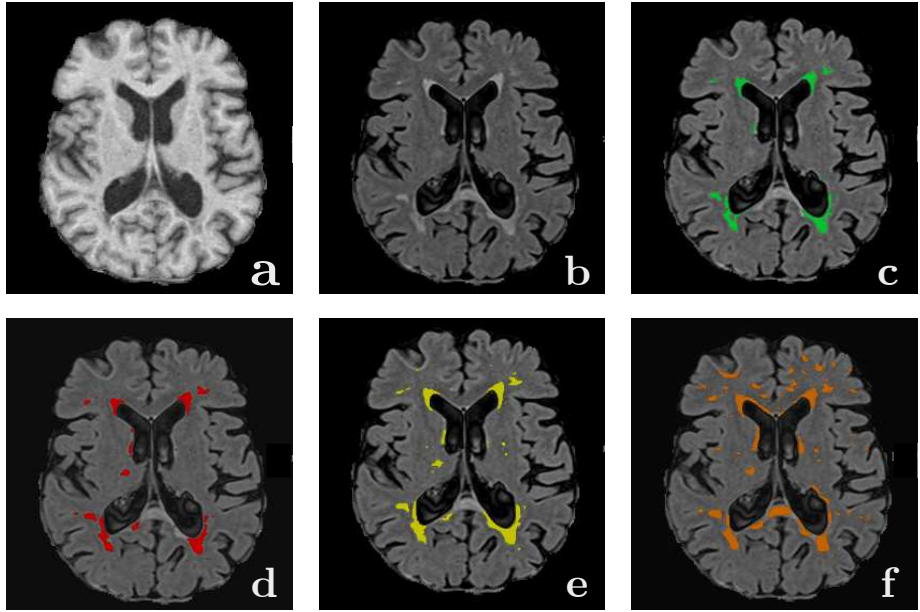}}
\caption{
Examples of the brain WMH MRI segmentations.
\textbf{(a)} Axial T1-weighted image. 
\textbf{(b)} FLAIR image.
\textbf{(c-f)} FLAIR images with WMH segmented labels:
\textbf{(c)} reference (green) WMH.
\textbf{(d)} WMH (red) from a domain adapted model ($\tilde{f}_{ST}(.)$) fine-tuned on five target training samples. 
\textbf{(e)} WMH (yellow) from model trained from scratch ($\tilde{f}_{ST}(.)$) on 100 target training samples.
\textbf{(f)} WMH (orange) from model trained from scratch ($\tilde{f}_{ST}(.)$) on 5 target training samples.
}
\label{fig:brains}
\end{figure}

The model trained on the set of images from the source domain ($\tilde{f}_S$), achieved a Dice score of 0.76. The same model, without fine-tuning, failed on the target domain with a Dice score of 0.005. Figure \ref{fig:plots}(a) demonstrates and compares the Dice scores obtained with three domain-adapted models to a network trained from scratch on different target training set sizes.
Figure~\ref{fig:plots}(b) illustrates the target domain test set Dice scores as a function of target domain training set size and the number of abstract layers that were fine-tuned.
Figure~\ref{fig:brains} presents and compares qualitative results of WMH segmentation of several different models of a single sample slice.

\section{Discussion and Conclusions}
We observed that while $\tilde{f}_S$ demonstrated a decent performance on $D_S$, it totally failed on $D_T$. 
Although the same set of learned representations is expected to be useful for both as the two tasks are similar, the failure comes to no surprise as the distribution of the responses to these features are different. 
Observing the comparisons presented by Figure~\ref{fig:plots}(a), it turns out that given only a small set of training examples on $D_T$, the domain adapted model substantially outperforms the model trained from scratch with the same size of training data. 
For instance, given only two training images, $\tilde{f}_{ST}$ achieved a Dice score of 0.63 on a test set of 33 target domain test images, while $\tilde{f}_T$ resulted in a dice of 0.15.
As Figure~\ref{fig:plots}(b) suggests, with only a few $D_T$ training cases available, best results can be achieved by fine-tuning only the last dense layers, otherwise enormous number of parameters compared to the training sample size would result in over-fitting.
As soon as more training data becomes available, it makes more sense to fine-tune the shallower representations (e.g., the last convolutional layers).
It is also interesting to note that tuning the first few convolutional layers is rarely useful considering their domain-independent characteristics.

\medskip
\bibliographystyle{unsrt}
\bibliography{sample}

\end{document}